\documentclass[a4paper]{article}
\usepackage{spconf,amsmath,graphicx}
\usepackage{amsfonts}
\usepackage{epsfig}
\usepackage{subcaption}
\usepackage{mwe}
\usepackage{xcolor}
\usepackage{hyperref}
\usepackage{algorithm}
\usepackage{algorithmic}
\usepackage{multirow}
\usepackage{adjustbox}

\title{Multi-task Language Modeling for Improving Speech Recognition \\of Rare Words}
\name{Chao-Han Huck Yang$^{1, 2*}$\thanks{$^*$The work is done as an applied scientist intern at Amazon Alexa.}, Linda Liu$^{2}$, Ankur Gandhe$^{2}$, Yile Gu$^{2}$, Anirudh Raju$^{2}$, Denis Filimonov$^{2}$, Ivan Bulyko$^{2}$}
\name{%
\begin{tabular}{@{}c@{}}
Chao-Han Huck Yang$^{*1, 2}$\thanks{$^*$The work is done as an applied scientist intern at Amazon Alexa.} \qquad
Linda Liu$^{1}$ \qquad
Ankur Gandhe$^{1}$ \\
Yile Gu$^{1}$ \qquad
Anirudh Raju$^{1}$ \qquad
Denis Filimonov$^{1}$ \qquad
Ivan Bulyko $^{1}$ 
\end{tabular}}
\address{$^1$Amazon, USA \\ $^2$Georgia Institute of Technology, USA }
\begin{document}
\ninept
\maketitle
\begin{abstract}
End-to-end automatic speech recognition (ASR) systems are increasingly popular due to their relative architectural simplicity and competitive performance.
However, even though the average accuracy of these systems may be high, the performance on \emph{rare} content words often lags behind hybrid ASR systems.
To address this problem, second-pass rescoring is often applied leveraging upon language modeling (LM). In this paper, we propose a second-pass system with multi-task learning, utilizing semantic targets (such as intent and slot prediction) to improve speech recognition performance. We show that our rescoring model trained with these additional tasks outperforms the baseline rescoring model, trained with only the LM task, by 1.4\% on a general test and by 2.6\% on a rare word test set in terms of word-error-rate relative (WERR). Our best ASR system with multi-task LM shows 4.6\% WERR deduction compared with RNN Transducer only ASR baseline for rare words recognition. 
\end{abstract}

\section{Introduction}
\label{sec1}

End-to-end neural systems have become widely used to solve a variety of natural language processing tasks. 
Particularly, in automatic speech recognition (ASR), systems like RNN-Transducer (RNN-T)~\cite{graves2012sequence} and Listen-Attend-Spell (LAS)~\cite{chan2016las}, have received much attention.
End-to-end system gained popularity in ASR in part because of their competitive performance and in part because of their architectural simplicity: instead of a number of different subsystems that go into a hybrid DNN-HMM system~\cite{povey11kaldi} (an HMM, a lexicon, a language model, etc.), end-to-end systems \cite{graves2012sequence, chan2016las} promise a single model that can be trained on paired sequences of acoustic features and words.
However, despite the overall competitive performance, end-to-end systems have been shown to have a relatively low performance on \emph{rare} words compared to hybrid systems, e.g.,~\cite{povey11kaldi}.
We believe that this may be due to the fact that hybrid systems take advantage of external knowledge outside of transcribed audio, such as human expert knowledge (lexicon) and larger language models trained on the unpaired text.

Second-pass rescoring is often used in ASR, which allows the use of a larger language model than would be practical within a single-pass system. In this work, we focus on improving a language model for second-pass rescoring.
We target a specific application: an interactive voice assistant. In recent years this type of application has received much attention and widespread use, including systems such as Cortana, Siri, Alexa, Google Home, to name a few.
Because these systems need to take action in response to user commands, they are often paired with a natural language understanding (NLU) module, which performs intent detection and slot tagging.
In this work, we propose a multitask training of a language model that utilizes the intent and slots as additional targets at training time, and explore how this method impacts rare word accuracy of rescoring hypotheses generated by an RNN-T first-pass system.

The rest of this paper is organized as follows: in Section~\ref{section:related-work}, we review related work, then we describe details of the proposed model in Section~\ref{sec:sec2} and present experimental results in Section~\ref{sec:exp:content}. 
Finally, we conclude in Section~\ref{sec:conclusion} and outline future work.

\begin{figure*}[hbt!]
\begin{center}
   \includegraphics[width=0.90\linewidth]{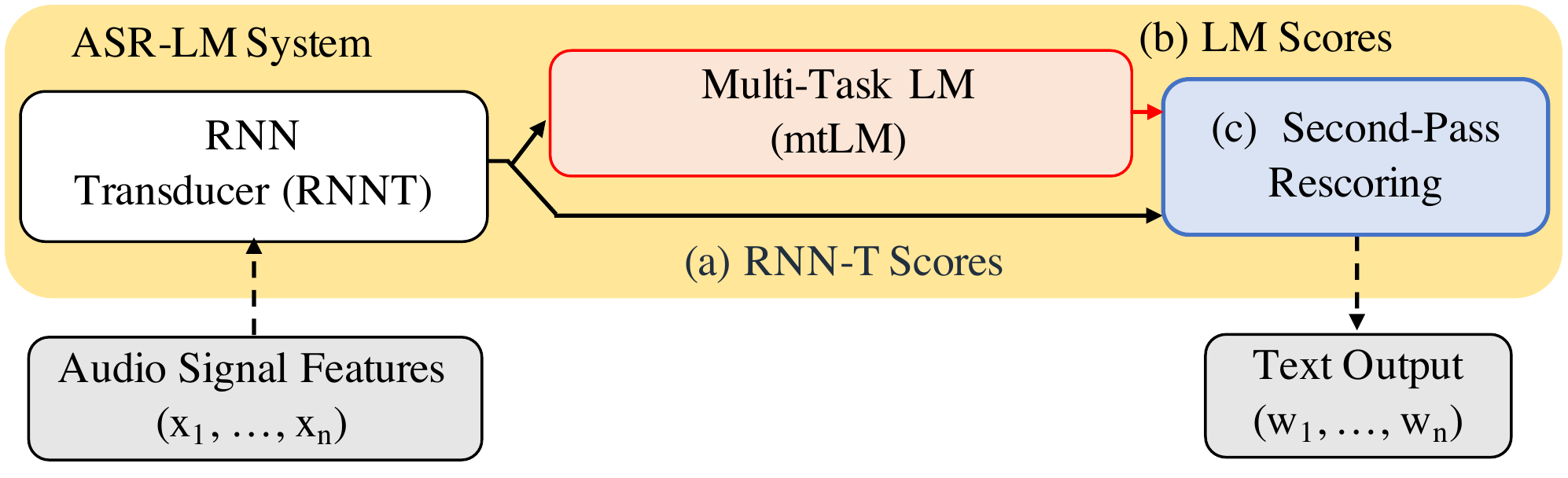}
\end{center}
   \caption{The proposed system includes (a) a first-pass RNN-Transducer and (b) a second-pass multi-task LM. The first and second-pass scores of the n-best ASR hypotheses are used for second-pass rescoring.
   } 
\label{fig:asr_block_diagram}
\end{figure*}

\section{Related Work}
\label{section:related-work}

Given the constrained search space in the second-pass,  second-pass rescoring models are able to leverage larger and more complex models than in the first-pass.
Previous works (e.g.,~\cite{raju2019scalable,sainath2019two,gandhe2020audio}) have shown that including the second-pass rescoring model can result in significant word error rate (WER) improvements.
Such second-pass rescoring systems are typically trained on various forms of textual data, including transcribed interactions with the ASR system, semi-supervised data, and other text corpora. More recently, 
several effective strategies have been identified for improving second-pass rescoring for end-to-end models,
including RNN-T. ~\cite{li2020towards, sainath2019two} used a first-pass RNN-T model combined with a LAS component for
second-pass rescoring and found that rescoring improved WER by up to 22\% relative. Similarly,
\cite{hu2020deliberation} used a deliberation model
for second-pass rescoring and found an additional 5-11\% relative WER reduction, compared to LAS rescoring.

The NLU module used in conjunction with the ASR system typically uses the outputs of the ASR system as its input to perform tasks like intent detection and slot tagging.
Intent detection refers to the task in which the ASR output (e.g., \emph{play the beatles}) is assigned an
intent (e.g., \emph{PlayMusicIntent}). Slot-filling refers to the task in which each word of the ASR output assigned a label (e.g., \emph{play$\vert$other
the$\vert$artist beatles$\vert$artist}). Numerous approaches, including support vector
machines~\cite{haffner2003optimizing} and neural networks~\cite{sarikaya2011deep}, have been used for
intent detection, a classification task. Popular approaches for slot-filling,
a sequence classification task, include conditional random fields~\cite{lafferty2001conditional} and recurrent neural networks
(RNNs)~\cite{yao2013recurrent}.

Typically, the models used to perform intent detection and slot-filling are trained separately. However, recent work showed that it is
possible to train a single model to perform  both of these tasks~\cite{guo2014joint,liu-lane-2016-joint}.~\cite{guo2014joint} 
demonstrated that the performance of a recursive neural network
jointly trained on intent (and domain) detection and slot-filling was comparable to that obtained by models trained on the
single-tasks alone. Similarly,~\cite{liu-lane-2016-joint} demonstrated that an attention-based RNN could
be trained to perform both tasks, resulting in an error reduction of 23.8\% on intents and 0.23\% on slots.
This and similar works~\cite{guo2014joint,xu2013convolutional} do not seek to improve ASR metrics like WER; instead, the ASR system is held fixed, and its outputs 
are used as inputs to the multi-task NLU module, and there is no ASR-related loss to backpropagate.

Given that ASR errors can affect downstream intent detection and slot-filling tasks (which would use the erroneous ASR as input),
~\cite{schumann2018incorporating, yang2020characterizing} incorporated an ASR error task into the attention-based RNN, trained to perform both intent detection and slot-filling, from
~\cite{liu-lane-2016-joint}. They found that including the ASR error task not only resulted in improvements for the
intent detection and slot-filling tasks but also significantly reduced the WER of the ASR engine by 4\% absolute.
Recently,~\cite{rao2020speech} proposed an end-to-end spoken language understanding system in which the ASR and NLU modules are combined -- the multi-task model showed improvements in both language understanding performance metrics, as well as ASR
performance metrics.  This suggests that information contained in the language understanding tasks can also benefit ASR.

Despite this, the focus of the majority of previous work on multi-task training with ASR and NLU tasks has typically been on the
benefits to NLU. In our work, we are instead interested in how signals from NLU can improve ASR,
specifically, with regards to language model performance. We demonstrate that a multi-task model trained with both NLU and ASR
objectives results in better WER when used as a second-pass rescoring model than a model trained with an ASR objective alone.

\section{Multi-task Language Modeling}
\label{sec:sec2}

\subsection{Neural language model}
\label{sec:nlm} 
We first define a standard neural language model (NLM) used in second-pass rescoring. Given a word sequence $\mathbf{w}$=($w_{1}, \ldots, w_{T}$), the likelihood of the word sequence can be written as: 
\begin{equation}
P(\mathbf{w})=\prod_{t=1}^{T+1} P\left(w_{t} \mid w_{< t}\right).
\end{equation}
where the sequence $\mathbf{w}$ is padded with additional symbols $w_{0}$ and $w_{T+1}$ to represent the   start-of-sentence ($<$sos$>$) and end-of-sentence ($<$eos$>$) tokens. 
Following a similar architecture  to ~\cite{raju2019scalable}, the model contains a word embedding layer followed by $K$ long short-term memory (LSTM)~\cite{sak2014long, hochreiter1997long} layers. For the $i$-th word, $P(\mathbf{w_i})$ is computed by taking a softmax over the unnormalized logits of all words, $l_i = exp(c^{T}_{K}e_{w_i}+b_i)$, where $c^{T}_{K}$ represents the hidden context vector of the $K^{th}$ layer and $e_{w_i}$ is the output word embedding vector and $b_i$ is the bias value. 
The training loss, $\mathbf{L_{LM}}$, is cross-entropy loss over $M$ samples: 
\begin{equation}
\mathbf{L_{LM}}=- \frac{1}{M} \sum_m \sum^{T+1}_{t=1} w_t^m \log p\left(w_{t}^m \mid w_{< t}; \theta_{LM}\right)
\label{eq:3}
\end{equation}

\subsection{Multi-task LM} 
\label{sec:multi_task_nlm} 
Our goal is to integrate the intent detection and slot filling sub-tasks into the training of our NLM and use this model for rescoring the output of the first pass, as shown in Figure~\ref{fig:asr_block_diagram}. We describe the shared architecture for each sub-task below: 

\textbf{Intent Detection} (ID) is a task of predicting the intent of a given sequence of words $\mathbf{w}$. In the multi-task setting with an additional language model task, the hidden representations $\mathbf{c_{K}}$ from the language model network (described in section~\ref{sec:nlm}) are fed into an intent detection specific network as shared features.  

The cross-entropy training objective for $\mathbf{L_{ID}}$ over $m$ samples is defined in Eq.~\ref{eq:4}):
\begin{equation} 
\mathbf{L_{ID}} =  - \frac{1}{m} \left [\mathit{log}(P(\mathbf{intent}\mid \mathbf{w}; \theta_{ID}, \theta_{LM}))\right ] ,
\label{eq:4}
\end{equation}
where $\theta_{LM}$ are the shared parameters between the LM task and ID task, and $\theta_{ID}$ are model parameters specific to intent detection.

\textbf{Slot Filling} (SF) is a task for predicting the slot label sequence $\mathbf{s}$ corresponding to the word sequence $\mathbf{w}$. The slot label sequence is represented as $\mathbf{s}$=($s_{1}, \ldots, s_{T}$), where $s_i$ is the slot label that aligns to input word $w_i$. Similar to intent detection, the hidden representations $\mathbf{c_{K}}$ from language model network are input to a slot filling network. The training objective, $\mathbf{L_{SF}}$  over $m$ samples is defined in Eq.~\ref{eq:5}:
\begin{equation} 
\mathbf{L_{SF}} = - \frac{1}{m} \left [\mathit{log}(P(\mathbf{s}\mid \mathbf{w}; \theta_{SF}, \theta_{LM}))\right ] , 
\label{eq:5}
\end{equation}
where $\theta_{SF}$ are the model parameters specific to slot filling network. 

\begin{figure*}[ht!]
\begin{center}
   \includegraphics[width=0.60\linewidth]{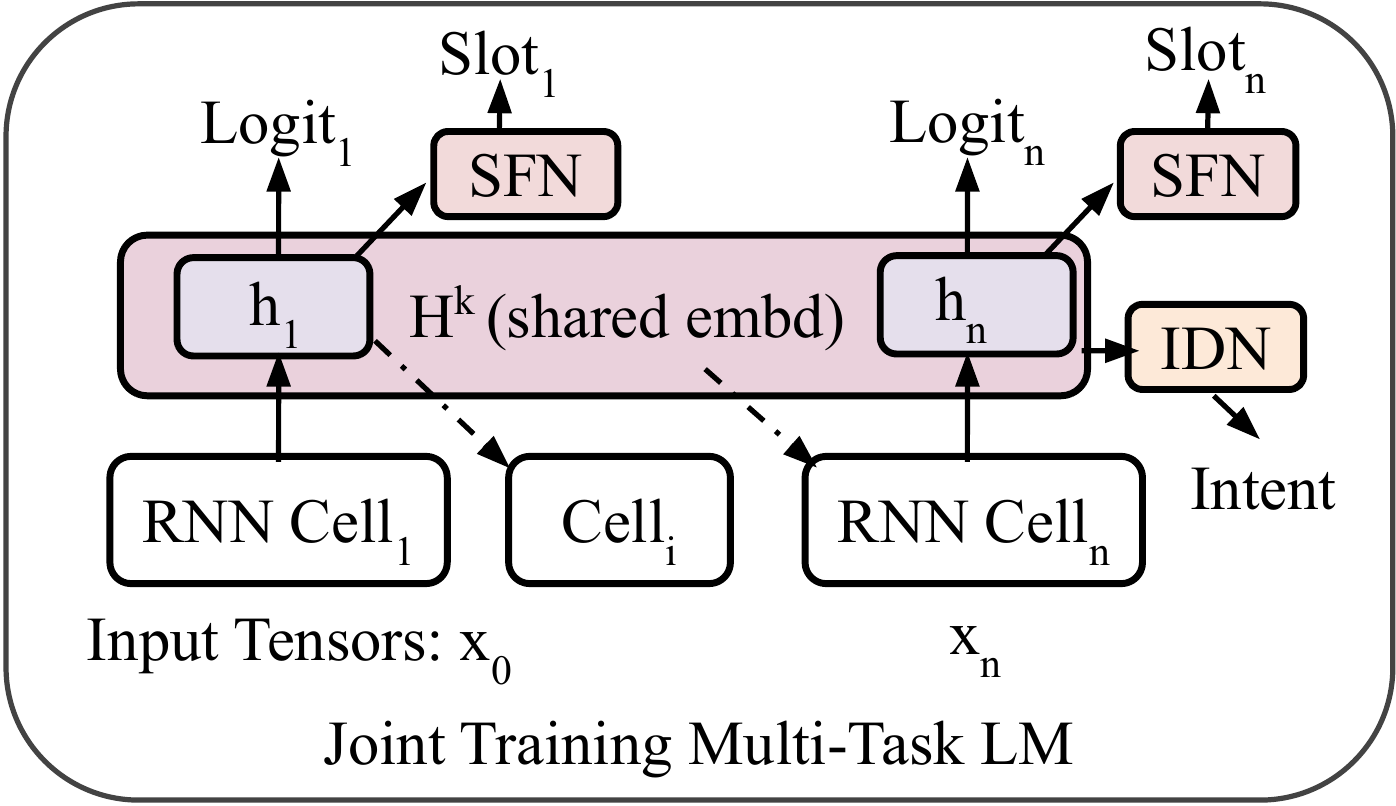}
\end{center} 
   \caption{A illustration of multi-task LM. Two sub-networks, which include slot filling network (SFN) and intent detection network (IDN), use the shared features from RNN.  
   } 
\label{fig:joint_training_model}
\end{figure*}

\subsection{Training objective for multi-task learning}
We introduce our training loss and optimization for our multi-task language model (\emph{mtLM}) shown as Fig.~\ref{fig:joint_training_model} in this section. The network is trained by gradient descent to find the parameters that minimize the cross-entropy of the predicted
and true distributions for the next word, intent class, and slot labels in a multi-task fashion. The total loss of mtLM, $\mathbf{L_{mtLM}}$, is defined by a normalized form:
\begin{equation}
\label{interpolation_mtLM}
    \mathbf{L_{mtLM}} = \alpha_{LM}\mathbf{L_{LM}}+\alpha_{ID}\mathbf{L_{ID}}+\alpha_{SF}\mathbf{L_{SF}},
\end{equation}
where $\alpha_{LM}$,$\alpha_{ID}$, and $\alpha_{SF}$ are weight coefficients for each sub-tasks in the \emph{mtLM}  corresponding
to Eq.~(\ref{eq:3}) to Eq.~(\ref{eq:4}). \\

To optimize these weight coefficients, we use the linear searching method~\cite{liu-lane-2016-joint} as the baseline in which the intent and slot's classification coefficients are set to zero from the beginning and linearly increased to one. Alternatively, we also apply the randomized weight majority algorithm (RWMA) from ~\cite{littlestone1994weighted, yang2020enhanced}. Compared to other gradient-based meta-learning methods, RWMA requires a lower model capacity and avoids executive computation while still smoothing the multi-task loss, as shown in ~\cite{chan1997accuracy, ren2018learning}.

\begin{algorithm}[tb!]
   \caption{Randomized Weighted Majority Algorithm (RWMA) on Multi-Task LM Training}
   \label{alg:wma}
\begin{algorithmic}
  \STATE {\bfseries 1.} \textbf{Initialize}: $\bar{\mathbf{\alpha}}_0 = (1, 1, 1)$ and $l_0 = (0,0,0)$ \\
  \STATE {\bfseries 2.} \textbf{For} $t = 1, 2, ..., T$ \\
  \STATE {\bfseries 3.} \hspace{4mm} Set $\alpha_{t} = \bar{\alpha}_{t} / Z_{t}$, where $Z_{t} = \sum_{i} \bar{\alpha}^{(i)}_{t}$. \\
  \STATE {\bfseries 4.} \hspace{4mm} \textbf{Update loss} $\mathbf{L_{mtLM}}$($\alpha_{LM}$=$\alpha^1_{t}$,~$\alpha_{ID}$=$\alpha^2_{t}$,~ $\alpha_{SF}$=$\alpha^{3}_{t}$) \\
  \STATE {\bfseries 5.} \hspace{4mm} Compute the step-wise loss $\mathbf{L^{step}_{mtLM}}$ and $l_t$. \\
  \STATE {\bfseries 6.} \hspace{4mm} If $\rho$($\mathbf{L^{step}_{LM}}$,$\mathbf{L^{i}}$) $<0$ over $\sum_{i=t} ({i-9}$, ... ,~${i})~\&~t>10$:\\
  \STATE {\bfseries 7.} \hspace{8mm} \textbf{Update rule $\forall i$}, $\bar{\alpha}_{t+1}^{(i)} = \bar{\alpha}_{t}^{(i)}exp((1-\eta)l_{t}) $. \\
  \STATE {\bfseries 8.} \hspace{4mm} Else: \\
  \STATE {\bfseries 9.} \hspace{8mm} \textbf{Keep rule $\forall i$}, $\bar{\alpha}_{t+1}^{(i)} = _{t}^{(i)} $. \\
\end{algorithmic}
\end{algorithm}

\begin{table*}[hbt!]
\centering
\begin{tabular}{|l|l|c|}
\hline
Training Data & Training Loss                                                                           & PPL$_{norm}$ \\ \hline
D$^{\mathbf{train}}_{\mathbf{NLU}}$ & $\mathbf{L_{LM}}$                                                    & 1.0                    \\
D$^{\mathbf{train}}_{\mathbf{NLU}}$ & $\mathbf{L_{LM}}$ + $\mathbf{L_{ID}}$                                                    & 0.995                    \\
D$^{\mathbf{train}}_{\mathbf{NLU}}$ & $\mathbf{L_{LM}}$ + $\mathbf{L_{SF}}$                                                    & 0.988                    \\
D$^{\mathbf{train}}_{\mathbf{NLU}}$ & $\mathbf{L_{LM}}$ + $\mathbf{L_{ID}}$ + $\mathbf{L_{SF}}$            & 0.985          \\ 
D$^{\mathbf{train}}_{\mathbf{NLU}}$ & $\mathbf{L_{LM}}$ + $\mathbf{L_{ID}}$ + $\mathbf{L_{SF}}$     + RWMA       & \textbf{0.982}          \\ 
\hline
D$^{\mathbf{train}}_{\mathbf{trans}}$ &             $\mathbf{L_{LM}}$                                                         & 0.985                   \\
D$^{\mathbf{train}}_{\mathbf{trans}}$$\overset{\mathbf{}}{\rightarrow}$ D$^{\mathbf{train}}_{\mathbf{NLU}}$ &
    $\mathbf{L_{LM}}$ $\overset{\mathbf{}}{\rightarrow}$$\mathbf{L_{LM}}$+$\mathbf{L_{ID}}$+$\mathbf{L_{SF}}$+RMWA & \textbf{0.974}  \\ 
\hline
\end{tabular}
    \caption{Normalized perplexity (PPL$_{norm}$) evaluated on general domain data (D$^{\mathbf{test}}_{\mathbf{gen}}$).
    PPL$_{norm}$ is computed against the result from the first row. 
    All models in the upper section are trained with the same multi-label NLU corpora
    (D$^{\mathbf{train}}_{\mathbf{NLU}}$) but different training objectives (LM, ID, SF) are used; lower section shows
    that for an LM pretrained with large transcription-only corpora D$^{\mathbf{train}}_{\mathbf{trans}}$, finetuning
    using multi-task training with smaller corpus D$^{\mathbf{train}}_{\mathbf{NLU}}$ still improves PPL$_{norm}$.}
\label{tab:1:general}
\end{table*}

\section{Experiments}
\label{sec:exp:content}

\subsection{Datasets}
We use $\sim$23k hours of far-field English utterances for voice control for training data to train our first-pass RNN-T model ~\cite{guo2020efficient}. For our second-pass NLMs, we used two training
sets: (1) The more intent data (D$^{\mathbf{train}}_{\mathbf{NLU}}$) described in~\cite{rao2020speech}, with 11k hours of English sentences, each human-annotated for
intent and slot; for our intent detection and slot-filling tasks, we used a subset of 25 intents and 20 slot labels from this dataset, based on the utility expected for our language modeling task. (2) A large-scale transcription-only corpora (D$^{\mathbf{train}}_{\mathbf{trans}}$) with $\sim$ 90k anonymized English sentences. Transcription only data is several times more plentiful than data containing both intent and slot annotations; we use (2) to demonstrate that multi-task training, even compared to a language model leveraging several more times training data. We have two evaluation sets consisting of: (1) a multi-label NLU corpora
(D$^{\mathbf{test}}_{\mathbf{gen}}$) with $\sim$900k English sentences representing the general use case and (2) a rare
words dataset (D$^{\mathbf{test}}_{\mathbf{rare}}$) of $\sim$100 hours of utterances, where each utterance must contain a word in the long-tail distribution. All corpora contain
de-identified utterances from voice assistant interactions.

\subsection{Model architecture}
\label{sec:4:2}
The baseline NLM architecture~\cite{raju2019scalable} contains a word embedding matrix of dimension 512 and two LSTM
layers of dimension 512 hidden, which is used as the shared feature representation $c_K$ for all three tasks. For the LM
task, the shared representation is fed to a softmax layer predicting 10k wordpieces~\cite{mike2012}.
For the ID and SF tasks, we use two baseline architectures from  ~\cite{liu2016attention}: we use the
aligned hidden
representation input over time on $c_K$ (no attention) and weighted average
of the hidden states (with attention), to encode the sequence representation to
ID and SF outputs. 

We further investigate an advanced architecture, projected attention layer, on NLU tasks with
parallel-connected attention proposed by~\cite{stickland2019bert}. We adapt two projected attention layers followed by
layer-norm over a weighted average layer of the hidden states $c_K$ of single attention head as the projected attention
encoder used in ID and SF. All the encoders have been carefully fine-tuned with a similar model parameters.

\subsection{Multi-task training objective}
To find the optimal the weight coefficients between different loss functions for \emph{mtLM} as in equation~\ref{interpolation_mtLM}, we use linear searching method~\cite{liu-lane-2016-joint} as the baseline, and study the impact of randomized weight majority algorithm (RWMA)~\cite{littlestone1994weighted, yang2020enhanced}. The RWMA is described in Algorithm~\ref{alg:wma}. 

We first initialize a number of experts equal to the number of sub-tasks, $d=3$. We choose $T=50$,  representing the total times for step-wise evaluation per epoch. We use a theoretical randomized coefficient $\eta =\sqrt{2\log (d) / T}$ and expert weight $\mathbf{\alpha}\in R^+$ with a range limit of (0.2, 0.6). $l$ is the probability of each expert showing degradations in performance over 1000 steps. Our online optimization is to minimize $\mathbf{L}_{LM}$ loss, as shown in step 6 of Algorithm~\ref{alg:wma}, where the individual coefficients $\bar{\alpha}^{i}$ would be adjusted with a negative Pearson correlation coefficient ($\rho$).

\subsection{Second-pass Rescoring}
As shown in Figure~\ref{fig:asr_block_diagram}, the multi-task trained language model is used to rescore the n-best hypotheses of a first-pass ASR. In this paper, we rescore the output of an RNN-T model. The RNN-T encoder includes of five LSTM layers. Each layer has 1024 hidden units. The prediction network of RNN-T has two LSTM layers of 1024 units and an embedding layer of 512 units, following ~\cite{guo2020efficient}. 

The one-best sequence is obtained  by combining the scores of RNN-T with the language model scores:
\begin{equation}
y^{*}=\arg \max _{y}\left(\log P(y \mid x)/|w|+\lambda \log P_{L M}(y) \right).
\label{eq:7:rescore}
\end{equation}
For LM scores, we use length-normalized probabilities from the language model network.  We determine coefficient $\lambda$ using grid search over a development data set; it was fixed to $0.006$ for all reported experiments. 

\subsection{Results} 
\begin{table*}[ht!]
\centering
\begin{tabular}{|l|c|c|c|c|}
\hline
\multirow{2}{*}{LM Model} & \multicolumn{2}{c|}{D$^{\mathbf{test}}_{\mathbf{gen}}$}  & \multicolumn{2}{c|}{D$^{\mathbf{test}}_\mathbf{rare}$}   \\ \cline{2-5} 
                                      & PPL$_{norm}$ & WERR (w/ RNN-T)         & PPL$_{norm}$ & WERR (w/ RNN-T)            \\ \hline
No rescoring                                & -              & baseline         & -              & baseline       \\ \hline
D$^{\mathbf{train}}_{\mathbf{NLU}}$+ $\mathbf{L_{LM}}$                             & 1.0          & -0.75\%          & 1.0        & -1.0\%        \\
D$^{\mathbf{train}}_{\mathbf{NLU}}$+ $\mathbf{L_{mtLM}}$                        & 0.982          & -2.5\%         & 0.984       &  -2.2\%       \\ \hline
D$^{\mathbf{train}}_{\mathbf{trans}}$+ $\mathbf{L_{LM}}$                              & 0.985          &   -2.0\% & 0.980         & -2.0\%        \\
D$^{\mathbf{train}}_{\mathbf{trans}}$ $\overset{\mathbf{}}{\rightarrow}$D$^{\mathbf{train}}_{\mathbf{NLU}}$+$\mathbf{L_{mtLM}}$& \textbf{0.974} & \textbf{-3.4\%}  & \textbf{0.971} & \textbf{-4.6\%} \\ \hline
\end{tabular}
\caption{
Rescoring results on general domain NLU data (D$^{\mathbf{test}}_{\mathbf{gen}}$) and rare words test set
    (D$^{\mathbf{test}}_\mathbf{rare}$). PPL values are normalized using the second row. WER relative$^{1}$ (WERR) is based on the first row
    as the baseline. The single-task models are trained with the language modeling loss $\mathbf{L_{SF}}$ and the
    multi-task models (rows 3 and 5) are trained with the multi-task loss $\mathbf{L_{SF}}$ with RWMA. The final row shows the effect of multi-task training when fine-tuing of a model trained with a large transcription corpora D$^{\mathbf{train}}_{\mathbf{trans}}$. }
\label{tab:3:rare:wma}
\end{table*}

\subsubsection{Model selection for multi-task LM}

We explore two types of language models: the single-task LM (\emph{stLM}) which only uses $\mathbf{L_{LM}}$ for
training, and the multi-task LM (\emph{mtLM}) which is discussed in detail in  \ref{sec:multi_task_nlm} and can use two
or more of $\mathbf{L_{LM}}$, $\mathbf{L_{ID}}$, and $\mathbf{L_{SF}}$ for training.

By evaluating perplexity (PPL) on general domain NLU data (D$^{\mathbf{test}}_{\mathbf{gen}}$), we assess the benefit of
multi-task training with additional ID and SF loss functions on the LM intrinsically. Throughout the paper, we
report normalized PPL (PPL$_{norm}$), where PPL is normalized over a fixed baseline PPL for a given test set.
Thus a PPL$_{norm}\textless$1 means that the candidate model's PPL improves over that of the baseline model.
As shown in the first four rows of Table~\ref{tab:1:general}, incorporating
additional losses from ID and SF both improve PPL: we observe a PPL$_{norm}$ reduction from 1.0
to 0.995 when we incorporate the ID loss ($\mathbf{L_{ID}}$) and a PPL$_{norm}$ reduction from 1.0 to 0.988 when we
incorporate the SF loss ($\mathbf{L_{SF}}$). We observe an additional benefit from training with all three losses for LM
and NLU tasks, from 1.0 to 0.985. This suggests that NLU information can potentially benefit LM training. 
Additionally, the final row of the first section in Table~\ref{tab:1:general} shows that optimizing $\mathbf{L_{mtLM}}$ using the RWMA shows additional benefit over the linear searching method described in Section 3.3. 

The last two rows of Table~\ref{tab:1:general} show that the benefit of multi-task training holds even when the language
model is trained on a large amount of training data. Often only a subset of training data may have NLU annotations
available. We leveraged additional (8x more) training data by using a transcription-only corpora
D$^{\mathbf{train}}_{\mathbf{large}}$ to train a \emph{stLM}. The PPL resulting from this LM was comparable to the
previous \emph{mtLM}, trained with smaller data set. However, in the final row, we see that fine-tuning this \emph{stLM}
using multi-task training with a smaller corpus D$^{\mathbf{train}}_{\mathbf{NLU}}$ improves PPL$_{norm}$) from 0.985 to 0.974.
Thus, the benefit of multi-task training appears to hold even when we are able to leverage a large amount of training
data. In the remaining experiments, we use the $\mathbf{L_{LM}}$, $\mathbf{L_{ID}}$, and $\mathbf{L_{SF}}$ to train \emph{mtLM}.

\begin{table}[ht!]
\centering
\begin{tabular}{|l|l|l|c|c|}
\hline
    Encoder Architectures                       & PPL$_{norm}$$\downarrow$ & ER$_{norm}$$\downarrow$ & F1$_{norm}$$\uparrow$ \\ \hline
No Attention ~\cite{liu2016attention}  & 0.992                    & 1.0                            & 1.0                      \\
Weighted-attention~\cite{liu2016attention}     & 0.988                 & 0.975                            & 1.013                      \\
Projected-attention                                & \textbf{0.985}                   & \textbf{0.906}                            & \textbf{1.028}              \\ \hline
\end{tabular}
    \caption{Comparison of multi-task LMs trained with different encoders. Normalized PPL (PPL$_{norm}$), intent error
    rate (ER$_{norm}$), and F1 score of slot-filling (SF$_{norm}$) are evaluated on 
    D$^{\mathbf{test}}_{\mathbf{gen}}$. PPL$_{norm}$ is computed against row 1 in Table~\ref{tab:1:general}. ER$_{norm}$
    and SF$_{norm}$ are computed against the first row here.}
\label{tab:2:lm:encod}
\end{table}

Table~\ref{tab:2:lm:encod} explores the result of using different types of feature encoders in
multi-task training described in section \ref{sec:4:2}. We show results using encoder architectures using aligned inputs
with and without attention following ~\cite{liu2016attention}, as well as 
with projected-attention~\cite{stickland2019bert} with residual connections. We find that
the projected-attention encoder (shown in Tab.~\ref{tab:2:lm:encod}) has better performance on normalized PPL, intent
error rate, and slot prediction F1 score. Our final model thus uses the architecture with additional projected-attention layers.

\subsubsection{Impact on WER in second-pass rescoring}

Next, we assess the benefit of multi-task training by using the \emph{mtLM} as a second-pass rescoring model. In Table~\ref{tab:3:rare:wma}, we report both the normalized PPL and WER\footnote{The performance of the RNN-T speech recognition baseline system presented in this work is below 10\% WER absolute.} relative (WERR) on both the general test set (D$^{\mathbf{test}}_{\mathbf{gen}}$) and rare words test set (D$^{\mathbf{test}}_\mathbf{rare}$). The WERR is measured relative to the first-pass RNN-T baseline (no rescoring). As before, the PPL values are normalized using the result from the single task LM trained with D$^{\mathbf{train}}_{\mathbf{NLU}}$ (second row). All multi-task LM results (rows 3 and 5) use the RWMA to optimize the multi-task loss, $\mathbf{L_{mtLM}}$.

First, we note that the baseline PPL for D$^{\mathbf{test}}_\mathbf{rare}$ is
around 3.3 times that of D$^{\mathbf{test}}_{\mathbf{gen}}$, and the baseline WER is roughly 2.5 times higher for
D$^{\mathbf{test}}_\mathbf{rare}$ (not shown in table). This is expected as D$^{\mathbf{test}}_\mathbf{rare}$ was
selected to contain utterances with at least one slot content word in the long-tail distribution, while
D$^{\mathbf{test}}_{\mathbf{gen}}$ represents the general use case. Consequently, we also expect that the impact of
multi-task training may be stronger for D$^{\mathbf{test}}_\mathbf{rare}$, given the presence of high WER slot content.
Next, comparing the single task LM D$^{\mathbf{train}}_{\mathbf{NLU}}$+ $\mathbf{L_{LM}}$ and multi-task LM
D$^{\mathbf{train}}_{\mathbf{NLU}}$+ $\mathbf{L_{mtLM}}$  trained on the smaller NLU dataset, we observe that multi-task
training results in an improvement of 1.75\% WERR on D$^{\mathbf{test}}_{\mathbf{gen}}$ and 1.2\% WERR on
D$^{\mathbf{test}}_\mathbf{rare}$ over the single-task model. We reported per-domain (intent) WER results in the rare word D$^{\mathbf{test}}_\mathbf{rare}$ in Table~4, where the Automation, Video, and Music domains show major improvements compared with the baseline of ASR-LM without the proposed multi-tasking learning.  

Critically, the same improvement trend is observed when using a single-task LM that is trained using all large
transcription-only corpora D$^{\mathbf{train}}_{\mathbf{trans}}$. In the final two rows of Table~\ref{tab:3:rare:wma},
we show that the model pre-trained on D$^{\mathbf{train}}_{\mathbf{trans}}$ and fine-tuned using multi-task training
with a smaller corpus performs better than the single-task model alone. The fact that multi-task LM can improve on
pretrained single-task LM is important, as we typically have access to large amounts of unannotated data and a small
amount of NLU labeled data. When having access to large transcription-only corpora
D$^{\mathbf{train}}_{\mathbf{large}}$, one can see that the improvement due to multi-task training is larger for the
rare words test set (-2.0\% to -4.6\% WERR) than the general test set (-2.0\% to -3.4\% WERR), underscoring the ability of multi-task LM to improve recognition on rare words. 
\\

\begin{table}[ht!]
\begin{adjustbox}{width=0.48\textwidth}
\begin{tabular}{|l|ccccc|}
\hline
Domain  & Music & Global & Notification & Automation & Video \\ \hline
WERR & -3.7\%       &  -2.4\%      &  -3.4\%            &   -4.8\%         &   -3.8\% \\ \hline
\end{tabular}
\end{adjustbox}
\caption{Per-domain WER relative$^{1}$ (WERR) improvement of RNN-T with multi-task LM ( $\mathbf{L_{mtLM}}$) compared with RNN-T with LM ($\mathbf{L_{LM}}$) on top-5 domains (intents) of the rare words dataset D$^{\mathbf{test}}_{\mathbf{rare}}$.}
\label{tab:4:in}
\end{table}

\section{Conclusion}
\label{sec:conclusion}
We demonstrate that incorporating the tasks of intent detection and slot-filling into language model training can reduce
perplexity and WER for ASR systems when the language model is used as a second pass rescoring model. We find that the improvement in WER is more pronounced for rare words, likely due to improvements in recognition of slot content. The proposed rescoring framework also shows good potential to advance existing RNN Transfer systems for domain-depended or customized continuous speech recognition~\cite{liu2021domain}. 

Moreover, we find the benefit of multi-task training holds even when comparing against a language model trained with large amounts of transcription data. Weighted optimization method~\cite{littlestone1994weighted, yang2020enhanced} is found to be a effective approach to determine the interpolation weights for the multi-task loss function for RNN Transducer modeling on rare word recognition. 

Future work aims to utilize the slot and intent predictions that are output from our multi-task model. For example, this can be done by modifying the decoder to consider paths conditioned not only on only previous words but also on previous slot labels. Additionally, we can use the final intent prediction to bias the second-pass rescoring results. The privacy-preserving~\cite{yang21_interspeech, novotney2021adjunct} learning is also an important future direction considering domain-specific speech and acoustic processing applications. For example, when the language transcription comes from a different or low-resource domain~\cite{pmlr-v139-yang21j} than the original speech corpora, it is necessary to provide reliable frameworks to measurement the information budgets of each training dataset. On the other hand, adaptation and knowledge distillation~\cite{novotney2021adjunct} are showing potential to get benefited from the multi-task learning techniques.

Finally, additional work can augment training with NLU data that is automatically labeled by intent or slot taggers to target further WER improvements. There are some recent works~\cite{thomas2021rnn, huang2020leveraging} discovering the connections between RNN Transducer modeling and spoken languages understanding. We believe that the effeteness of multi-task learning discussed in this work could be useful to design large-scale pre-training methods for more down-stream applications.

\clearpage
\bibliographystyle{IEEEbib}
\bibliography{refs}

\end{document}